\PassOptionsToPackage{table,dvipsnames}{xcolor}
\documentclass[sigconf]{acmart}

\usepackage{epsf}
\usepackage{graphicx}
\usepackage{hyperref}
\usepackage[ruled,vlined]{algorithm2e}
\usepackage{booktabs, makecell, tabularx}
\usepackage{amsmath}
\usepackage{mathtools}
\usepackage{color}
\usepackage[table,dvipsnames]{xcolor}    
\usepackage{comment}
\usepackage[labelsep=period]{caption}
\usepackage{soul} 
\usepackage{booktabs} 
\usepackage[normalem]{ulem} 
\usepackage{multirow}
\usepackage{color}
\usepackage{subcaption}
\usepackage{booktabs, makecell, tabularx}
\usepackage{multirow}
\newcolumntype{?}{!{\vrule width 1pt}}

\makeatletter
\def\hlinewd#1{%
\noalign{\ifnum0=`}\fi\hrule \@height #1 %
\futurelet\reserved@a\@xhline}
\makeatother

\begin{document}
\title[A Novel Surrogate-assisted EA Applied to Partition-based Ensemble Learning]{A Novel Surrogate-assisted Evolutionary Algorithm Applied to Partition-based Ensemble Learning}

\author{Arkadiy Dushatskiy}
\affiliation{%
  \institution{Centrum Wiskunde \& Informatica}
  \city{Amsterdam} 
  \state{the Netherlands} 
}
\email{arkadiy.dushatskiy@cwi.nl}

\author{Tanja Alderliesten}
\affiliation{%
  \institution{Leiden University Medical Center}
  \city{Leiden} 
  \state{the Netherlands} 
}
\email{t.alderliesten@lumc.nl}

\author{Peter A. N. Bosman}
\affiliation{%
  \institution{Centrum Wiskunde \& Informatica}
  \city{Amsterdam} 
  \state{the Netherlands} 
}
\affiliation{%
  \institution{TU Delft}
  \city{Delft} 
  \state{The Netherlands} 
}
\email{peter.bosman@cwi.nl}

\begin{abstract}
We propose a novel surrogate-assisted Evolutionary Algorithm for solving expensive combinatorial optimization problems. We integrate a surrogate model, which is used for fitness value estimation, into a state-of-the-art P3-like variant of the Gene-Pool Optimal Mixing Algorithm (GOMEA) and adapt the resulting algorithm for solving non-binary combinatorial problems. We test the proposed algorithm on an ensemble learning problem. Ensembling several models is a common Machine Learning technique to achieve better performance. We consider ensembles of several models trained on disjoint subsets of a dataset. Finding the best dataset partitioning is naturally a combinatorial non-binary optimization problem. Fitness function evaluations can be extremely expensive if complex models, such as Deep Neural Networks, are used as learners in an ensemble. Therefore, the number of fitness function evaluations is typically limited, necessitating expensive optimization techniques. In our experiments we use five classification datasets from the OpenML-CC18 benchmark and Support-vector Machines as learners in an ensemble. The proposed algorithm demonstrates better performance than alternative approaches, including Bayesian optimization algorithms. It manages to find better solutions using just several thousand fitness function evaluations for an ensemble learning problem with up to 500 variables.
\end{abstract}


\maketitle

\section{Introduction}
Expensive combinatorial optimization is a combinatorial optimization subfield, in which fitness function evaluations are (computationally) expensive. In contrast to the case of optimization problems with inexpensive function evaluations, in case of problems with expensive function evaluations, search algorithms are usually evaluated by their ability to find good solutions within a limited number of function evaluations rather than by a required number of evaluations to find the global optimum. Expensive optimization problems arise in various domains. One traditional application is simulation-based optimization in engineering design \cite{kleijnen2010constrained, koziel2015efficient, rashid2013efficient}. Combinatorial expensive optimization has recently achieved increased attention due to the popularity of the Neural Architecture Search (NAS) field \cite{elsken2019neural, ottelander2020local}. The goal in NAS is to find the best Neural Network architecture for a particular task. A function evaluation that corresponds to a partial \cite{bender2018understanding} or end-to-end \cite{real2019regularized} Deep Neural Network (DNN) training procedure can be extremely expensive, taking up to several hours \cite{Ghiasi_2019_CVPR}. A related application is finding an ensemble of deep learning models through data partitioning, for instance to tackle the problem of data heterogeneity in medical image analysis \cite{dushatskiy2020observer, mirikharaji2020d}. Here too, the efficiency of the search algorithm is extremely important as each function evaluation again requires training a DNN. Therefore, novel efficient algorithms for combinatorial expensive optimization are highly demanded.

Bayesian Optimization (BO) is a traditional approach to solving expensive optimization problems. The most common BO algorithms are based on Gaussian processes, aimed at real-valued variables. For their use in the combinatorial domain, they were shown to have caveats \cite{bointeger}. Consequently, novel BO algorithms were developed, designed to handle categorical variables. Replacing the Gaussian Process surrogate model with Random Forests \cite{breiman2001random} was proposed in Sequential Model-based Algorithm Configuration (SMAC) \cite{smac}. Another popular BO algorithm, which was shown to perform better than traditional Gaussian process-based BO in categorical and mixed domains, is the Tree-structured Parzen Estimator (TPE) \cite{tpe} and its implementation in the Hyperopt package \cite{hyperopt}. Recently, two BO algorithms with better performance than SMAC and Hyperopt were proposed: Bayesian Optimization of Combinatorial Structures (BOCS) \cite{bocs} and Combinatorial Bayesian Optimization using the Graph Cartesian Product (COMBO) \cite{combo}. The main idea of both these algorithms is to explicitly model interactions between variables. While BOCS takes into account only second-order interactions, COMBO has no limit on the order of interactions and builds a sparse Bayesian regression model over possible high-order interactions between variables. BOCS is limited to binary variables only, while COMBO can handle variables with higher cardinality. COMBO showed state-of-the-art performance on both common binary optimization benchmark problems such as Ising Spin-glass and MAXSAT and an example of NAS. However, this comes at a cost of extremely expensive computations that grows fast as the number of function evaluations increases for maintaining the surrogate model. For instance, only experiments with up to 300 function evaluations were feasible in the original paper \cite{combo}.

Surrogate-assisted Evolutionary Algorithms (EAs) are an interesting alternative approach to solving expensive optimization problems. Using surrogate models to replace part of the function evaluations done by an EA is a natural way to reduce the number of evaluations while keeping the powerful search characteristics of EAs. Common surrogate models are polynomials, Kriging, Radial Basis Function Network (RBFN), or Support Vector Regression (SVR) \cite{surrogate_ea_review}. However, surrogate-assisted EAs are not as widely used for combinatorial optimization problems as for real-valued ones. A recent approach for combinatorial optimization integrates a Convolutional Neural Network with an advanced model-based EA called Convolutional Neural Network Surrogate-assisted GOMEA (CS-GOMEA), but it is limited to binary problems \cite{dushatskiy2019convolutional}.

Local Search (LS) is a well-known simple, yet often effective search algorithm. The main principle is to perform greedy search (only improvements are accepted) in the vicinity of a current solution, leading to quick convergence to a local optimum. It was recently shown that LS can be very efficient for combinatorial formulations of NAS, achieving competitive performance with state-of-the-art EAs and outperforming a commonly adopted baseline in NAS - Random Search (RS)~\cite{ottelander2020local, white2020local}.

Ensembling is a common Machine Learning technique to improve performance. Usually, individual models are simple learners such as Decision Trees. A classic approach is Bagging \cite{breiman1996bagging}, where several models are trained on randomly selected subsets of samples. An alternative approach is to train models on disjoint subsets of samples \cite{alpaydin1997voting, chawla2004learning}. In contrast to Bagging, with such an approach it is possible to train each model on homogeneous data, increasing chances of better performance. Finding the best partitioning then, however, is a potentially expensive optimization problem.

In this paper, we propose a novel surrogate-assisted EA based on a state-of-the-art P3-like variant of GOMEA. Unlike CS-GOMEA \cite{dushatskiy2019convolutional}, it is not limited to binary problems, does not make assumptions about problem regularity (subfunctions of fixed size), and uses a more efficient population management scheme - Parameterless Population Pyramid (P3). To the best of our knowledge, this is the first time a surrogate model is integrated into a P3-like EA. We also propose a simple, yet shown to be efficient adaptive mechanism to control and balance the number of performed real and surrogate fitness evaluations. The introduced algorithm is applied to an ensembling problem defined on supervised classification datasets. Support-vector Machines (SVMs) are used as learners in the considered ensembles.

\section{Search problems and algorithms}

\subsection{Problem formulation}
In general, we consider unconstrained combinatorial optimization problems with categorical variables. Potentially, the cardinality of problem variables differs. However, for simplicity of notation in this work, we assume that all variables have cardinality $\alpha$, meaning that all variables have $\alpha$ possible values. Mathematically, the considered problems can be formulated as a global optimization problem $\max_{x \in \mathcal{D}}f(x)$ where $f(x):\mathcal{D} \rightarrow \mathbb{R}$, $\mathcal{D}=\{0,\dots,\alpha-1\}^\ell$, 
$\ell$ is the number of variables, and $\alpha$ is the alphabet size.

\subsection{Local Search}
In contrast to RS, LS traverses the search space in an ordered, local manner, quickly converging to a local optimum. In a random restart scenario, LS starts from a randomly generated solution. We specifically consider first improvement neighbourhood search as our LS. A single iteration of LS consists of considering all variables in a random order, and assigning to each variable the value in its domain that corresponds the best fitness function value. Iterations are repeated until a solution is not changed. After that, a new random solution is generated and the search loop is repeated.
The Random Restart Local Search pseudocode is provided in Algorithm~\ref{alg:LS}.

 \def\HiLi{\leavevmode\rlap{\hbox to \hsize{\color{yellow!50}\leaders\hrule height .8\baselineskip depth .5ex\hfill}}}
 
\begin{algorithm}
 \DontPrintSemicolon
\footnotesize
 \SetKwFunction{randomPermutation}{randomPermutation}
 \SetKwFunction{evaluate}{evaluate}
 \SetKwFunction{estimate}{estimate}

 \SetKwFunction{randomlyInitialize}{randomlyInitialize}
 \SetKwFunction{generateRandomSolution}{generateRandomSolution}
 \SetKwFunction{cp}{copy}
 \SetKwFunction{termination}{TerminationCriterionSatisfied}
 \SetKwFunction{collectRandomSolutions}{collectRandomSolutions}
 \SetKwFunction{trainSurrogateModel}{trainSurrogateModel}

 \SetKwFunction{LS}{LocalSearch}
 \SetKwFunction{SALS}{SurrogateAssistedLocalSearch}
\SetKwProg{Fn}{Function}{:}{}
\SetKwData{true}{true}
\SetKwData{false}{false}
\SetKwRepeat{Do}{do}{while}
\SetKwInOut{Parameter}{parameter}

\Fn{\LS{}}
{
\Do{\text{budget is not exhausted}}
{
 $improved = \false$ \;
 $s \gets \text{\generateRandomSolution{}}$ \;
\Do{$improved$}
{  
 $s.fitness \gets$ \evaluate{$s$}\;
 \For{$i \in$ \randomPermutation{$\{0, \dots, \ell-1\}$}}
 {
     $s^\prime \gets s$ \;
     \For{$j \in \{0,\dots,\alpha-1\} \setminus s_i$}
     {
     $ s^{\prime}_i \gets j $\;
     ${s^\prime}.fitness \gets$ \evaluate{$ s^\prime $} \;
     \If{${s^\prime}.fitness > s.fitness$}
     {
         $s_i \gets s^{\prime}_i$\;
         $improved \gets \true$ \;
     }
     }
 }
}
}
}

\caption{Random Restart Local Search Algorithm.}
\label{alg:LS}
\end{algorithm}

\subsection{An adaptation of P3 for non-binary problems}
We adapt the state-of-the-art model-based Evolutionary Algorithm known as the Parameterless Population Pyramid (P3) \cite{goldman2015fast} for solving non-binary combinatorial optimization problems. P3 is a variant of the Gene-pool Optimal Mixing Algorithm (GOMEA) \cite{bosman2012linkage} that uses local search, a complementary donor search, and a fitness-pyramid structured growing population. We call the P3 variant used in this paper \emph{P3-GOMEA}. Below, we give a brief overview of its features. Pseudocode is provided in Algorithm~\ref{alg:P3} and Algorithm~\ref{alg:P3Func}.

\subsubsection{Linkage Model}
The idea of using a so-called Linkage Model for guiding the evolution process through variation restricted to specific variables and immediately testing for improvements was introduced with the concept of Optimal Mixing Evolutionary Algorithms (OMEAs) \cite{thierens2011optimal}. A Linkage Model is commonly defined in the form of a set of (possibly overlapping) subsets of variables. Such a structure is called a Family Of Subsets (FOS). The goal is to use information about dependencies between variables to perform crossover more efficiently. In the case of Black-Box optimization these dependencies are not known a priori, and, therefore, they have to be learned during optimization. A much used approach is to first estimate pairwise dependencies from the population and build higher-order models upon this. For pairwise dependencies learning either Mutual Information (MI) \cite{thierens2011optimal} or Normalized Mutual Information (NMI) \cite{goldman2015fast} is often used. Both MI and NMI calculations can be naturally extended to the case of non-binary variables \cite{luong2015scalable}. P3-GOMEA uses the NMI measure. After pairwise dependencies are learned, they can be used for Linkage Model construction.

The FOS model known as the Linkage Tree (LT), which uses a hierarchical structure to store sets of dependent variables, was shown to often be the most efficient one \cite{thierens2011optimal}. An LT is a binary tree with $2\ell-1$ nodes. LT leaves are singletons with variables. The root of an LT is a set containing all variables. All other nodes are linkage subsets $F^i$ which are unions of disjoint subsets of children $k,j$ of node $i$: $F^i=F^j \cup F^k, F^j \cap F^k = \emptyset$. 

P3-GOMEA uses a filtered LT Model as proposed in \cite{bosman2013more}, in which redundant subsets are removed. When two subsets $F^j$ and $F^k$ are merged into a subset $F^i$, it may happen that the MI or NMI measure between them is maximal (one). It is supposed that there is then no merit in using these subsets in variation separately as it may disrupt a building block $F^i$. Thus, keeping subsets $F^j$ and $F^k$ in an FOS is not reasonable and only the parent subset $F^i$ is kept. 

\subsubsection{Gene-pool Optimal Mixing}
The Gene-pool Optimal Mixing operator (GOM) is the variation operator used in the GOMEA family of algorithms. GOM is applied to a single solution and outputs a single solution that is never worse than the input solution. To improve a solution, GOM loops over the contents of the FOS model. 
For each linkage subset $F^i$, GOM attempts to overwrite the values of the variables in $F^i$ of the solution in consideration, with values from a donor solution, often chosen at random from the population.
If this overwriting action does not cause the fitness of the solution to become
worse, the copy action is accepted. Otherwise, the donor material is rejected
and the action is undone. 
The pseudocode of GOM is provided in Algorithm~\ref{alg:P3Func}.

\subsubsection{Parameterless population management scheme}
Choosing an optimal population size is not trivial, and, therefore, a population management scheme without a population size parameter has much practical value. Several such schemes were proposed for the GOMEA family of algorithms \cite{den2016multiple}. In this paper, we consider P3 \cite{goldman2015fast}, which was shown to be efficient \cite{den2016multiple}, is fully parameterless, and can be naturally adapted to a surrogate-assisted EA. 

Solutions are stored in a pyramidal structure. Each layer of the pyramid is a set of solutions (duplicates are not stored). Linkage Models are learned separately for each pyramid level.
In each iteration, only one solution is evolved. A new solution is generated randomly and added to the bottom layer of the pyramid. Then, by using solutions from the current pyramid level as donors, the current solution is evolved using GOM. If GOM leads to fitness improvement, the resulting solution is added to the next pyramid level, and GOM is performed at the next pyramid level. Otherwise, processing this solution is finished, and a new iteration starts with a new, randomly generated solution.

\subsubsection{Hill Climber}
Before evolving a solution using GOM, an LS algorithm (also called a Hill Climber in \cite{goldman2015fast}) can be applied to quickly bring the solution to a local optimum. Such an approach is used in the original P3 algorithm \cite{goldman2015fast}. However, our preliminary experiments showed that in the case of a limited number of allowable function evaluations (several thousand evaluations) such use of LS leads to similar performance as stand-alone LS (without P3) because the majority of function evaluations are spent on the LS phase. Therefore, we do not include LS in our P3-GOMEA.  

\begin{algorithm}
\SetAlFnt{\tiny}

 \def\HiLi{\leavevmode\rlap{\hbox to \hsize{\color{yellow!50}\leaders\hrule height .8\baselineskip depth .5ex\hfill}}}
\def\HiGr{\leavevmode\rlap{\hbox to \hsize{\color{gray!50}\leaders\hrule height .8\baselineskip depth .5ex\hfill}}}

 \DontPrintSemicolon
\footnotesize
 \SetKwFunction{SurrogateAssistedPThree}{SurrogateAssistedP3}
 \SetKwFunction{growthFunction}{growthFunction}
 \SetKwFunction{createRandomSolution}{createRandomSolution}
 \SetKwFunction{generateRandomSolutions}{generateRandomSolutions}
 \SetKwFunction{hillClimber}{hillClimber}
 \SetKwFunction{learnModel}{learnLinkageModel}
 \SetKwFunction{trainSurrogateModel}{trainSurrogateModel}
 \SetKwFunction{predictSurrogateFitness}{predictSurrogateFitness}
 \SetKwFunction{compareSolutions}{compareSolutions}
 \SetKwFunction{setThreshold}{setThreshold}
 \SetKwFunction{GOM}{GOM}
 \SetKwFunction{EA}{EA}
 \SetKwFunction{evaluateSolution}{evaluateSolution}
 \SetKwFunction{acceptChange}{acceptChange}
 \SetKwFunction{calculateFitness}{calculateFitness}
 \SetKwFunction{predictSurrogateFitness}{predictSurrogateFitness}
 \SetKwInOut{Parameter}{Parameters}
\SetKwProg{Fn}{Function}{:}{}
\SetKwData{true}{true}
\SetKwData{false}{false}
\newcommand\mycommfont[1]{\footnotesize\ttfamily\textcolor{blue}{#1}}
\SetCommentSty{mycommfont}

\Parameter{A relaxation parameter $\eta$}
\Fn{\EA{}}
{
$iter \gets 0$ \;
$Pyramid \gets [\emptyset$] \;
\HiLi $R \gets \emptyset$ \;
\HiLi \For{$i=0,\dots,\ell-1$}
{
    \HiLi $R \gets R \cup \generateRandomSolution{}$ \;
}
\HiLi $\trainSurrogateModel(R)$ \;
\HiLi $F \gets \predictSurrogateFitness(R) $ \;
\HiLi $\lambda \gets 1$ \;
\HiLi $\mathcal{T} \gets$ \setThreshold{$F, \lambda$} \;

\While{$\neg terminationCriterionSatisfied$}
    {
        $elitist \gets None$ \tcp{elitist value is updated if necessary when real evaluations are performed}
        $p \gets \generateRandomSolution{}$ \;
        $Pyramid^0 \gets Pyramid^0 \cup \{p\}$ \;
        $solutionsAdded \gets \true$ \;
        $currentTopLevel \gets |Pyramid|-1$ \;
        $\mathcal{L} \gets 0$ \;
        $elitistBefore \gets elitist$ \;
        \While {$\mathcal{L} \le currentTopLevel$ and $solutionsAdded$}
        {
            $\mathcal{F} \gets \learnModel(Pyramid^\mathcal{L})$ \;
                $o \gets \GOM(p, Pyramid^\mathcal{L}, \mathcal{F})$ \;
                \If {\compareSolutions{$o, p$}}
                {
                    \If{$\mathcal{L}=currentTopLevel$}
                    {
                        $Pyramid^{\mathcal{L}+1}.append(\emptyset)$ \;
                    }
                    $Pyramid^{\mathcal{L}+1} \gets Pyramid^{\mathcal{L}+1} \cup \{o\}$ \;
                    $solutionsAdded \gets \true$ \;
                }
            $p \gets o$ \;
            $\mathcal{L} \gets \mathcal{L}+1$ \;
        }
        $elitistAfter \gets elitist$ \;
        \HiLi \If {$\neg \compareSolutions{elitistAfter, elitistBefore}$}
        {
            \HiLi $\lambda \gets \lambda \eta $ \;
        }
        \HiLi $\trainSurrogateModel{R}$ \;
        \HiLi $F \gets \predictSurrogateFitness(R) $ \;
        \HiLi $\mathcal{T} \gets$ \setThreshold{$F, \lambda$} \;
    }
}

\caption{P3-GOMEA and Surrogate-assisted P3-GOMEA (SA-P3-GOMEA). The lines which are added in the surrogate-assisted algorithm are highlighted in yellow.
\label{alg:P3}}

\end{algorithm}

\begin{algorithm}
\SetAlFnt{\tiny}

 \def\HiLi{\leavevmode\rlap{\hbox to \hsize{\color{yellow!50}\leaders\hrule height .8\baselineskip depth .5ex\hfill}}}
\def\HiGr{\leavevmode\rlap{\hbox to \hsize{\color{gray!50}\leaders\hrule height .8\baselineskip depth .5ex\hfill}}}

 \DontPrintSemicolon
\footnotesize

 \SetKwFunction{SurrogateAssistedPThree}{SurrogateAssistedP3}
 \SetKwFunction{growthFunction}{growthFunction}
 \SetKwFunction{createRandomSolution}{createRandomSolution}
 \SetKwFunction{generateRandomSolutions}{generateRandomSolutions}
 \SetKwFunction{hillClimber}{hillClimber}
 \SetKwFunction{learnModel}{learnLinkageModel}
 \SetKwFunction{trainSurrogateModel}{trainSurrogateModel}
 \SetKwFunction{predictSurrogateFitness}{predictSurrogateFitness}
 \SetKwFunction{compareSolutions}{compareSolutions}
 \SetKwFunction{setThreshold}{setThreshold}
 \SetKwFunction{GOM}{GOM}
 \SetKwFunction{EA}{EA}
 \SetKwFunction{evaluateSolution}{evaluateSolution}
 \SetKwFunction{acceptChange}{acceptChange}
 \SetKwFunction{calculateFitness}{calculateFitness}
 \SetKwFunction{predictSurrogateFitness}{predictSurrogateFitness}
 \SetKwProg{Fn}{Function}{:}{}
\SetKwData{true}{true}
\SetKwData{false}{false}
\newcommand\mycommfont[1]{\footnotesize\ttfamily\textcolor{blue}{#1}}
\SetCommentSty{mycommfont}
\SetKwInOut{Input}{Variables}

\Input{Real elitist solution \textit{elitist}, threshold $\mathcal{T}$, quantile $\lambda$, set of solutions with known real fitness values $R$}

 \Fn{\GOM{o, $\mathcal{P}, \mathcal{F}$}}
{
$backup \gets o$ \;
$changed \gets \false$ \;
$\mathcal{F} \gets sortFOSInAscendingOrder(\mathcal{F})$ \;
\For{$i \in \{0,1,\dots,|\mathcal{F}|-1\}$}
    {
        $donorsList = \randomPermutation (\{\mathcal{P}_0, \mathcal{P}_1,\dots,\mathcal{P}_{n-1}\})$\;
        \For{$j \in \{0,1,\dots,n-1\}$}
        {
            $d \gets donorsList[j]$\;
            $o_{F^i} \gets d_{F^i}$\;
            \If {$o_{F^i} \ne d_{F^i}$}
                {
                    $\evaluateSolution(o)$\;
                    
                    \If{\acceptChange(o)}
                    {
                        $backup_{F^i} \gets o_{F^i}$\;
                        $changed \gets \true$ \;
                    }
                    \Else
                    {
                        $o_{F^i} \gets backup_{F^i}$\;
                    }

                    break\;
                }
        }   
    }
    return $o$ \;
}

\Fn{\evaluateSolution{x}}
{
\HiGr $x.fitness \gets \calculateFitness(x)$\;
 \HiGr $x.realFitnessCalculated \gets true$ \;

\HiLi $x.surrogateFitness \gets \predictSurrogateFitness(x)$\;
\HiLi \If {$x.surrogateFitness > \mathcal{T}$}
{
     \HiLi $x.fitness \gets \calculateFitness(x)$\;
     \HiLi $x.realFitnessCalculated \gets true$ \;
     \HiLi $R \gets R \cup \{x\}$ \;
}
\If{x.realFitnessCalculated}
{
    \If{x.fitness > elitist.fitness}
    {
    $elitist \gets x$ \;
    \HiLi $\lambda \gets 1$ \;
    }
}
}

\Fn{\compareSolutions{x,y}}
{
\HiLi \If{x.realFitnessCalculated and y.realFitnessCalculated}
{
\If{x.realFitness > y.realFitness}{return 1;}
\If{x.realFitness == y.realFitness}{return 0;}
\If{x.realFitness < y.realFitness}{return -1;}

}
\HiLi \Else
{
\HiLi \If{x.surrogateFitness > y.surrogateFitness}{\HiLi return 1;}
\HiLi \If{x.surrogateFitness == y.surrogateFitness}{\HiLi return 0;}
\HiLi \If{x.surrogateFitness < y.surrogateFitness}{\HiLi return -1;}
}
}

\caption{Key functions used in the P3-GOMEA and SA-P3-GOMEA algorithms. The lines of code which are not used in the surrogate-assisted algorithm are highlighted in gray. The lines which are added in the surrogate-assisted algorithm are highlighted in yellow.
\label{alg:P3Func}}
\end{algorithm}

\subsection{Surrogate-assisted EA}
We integrate, in a novel way, a surrogate model into the above-described
P3-GOMEA to replace a part of the real function evaluations with surrogate
function evaluations. Our proposed approach to integrating surrogate models can
be applied to various search algorithms. In this work, we focus on a
surrogate-assisted version of the P3-GOMEA algorithm that we refer to as
\emph{SA-P3-GOMEA}. The changes required to P3-GOMEA to obtain SA-P3-GOMEA are
provided in Algorithm~\ref{alg:P3} and Algorithm~\ref{alg:P3Func}.

\subsubsection{Surrogate model integration}
The outline of the proposed surrogate model integration into a search algorithm
is as follows. First, some initial random solutions are evaluated. In general,
the number of initial solutions can be tuned, but we use a simple approach that
is also used in COMBO, namely, $\ell$ random solutions are generated and
evaluated. Then, a surrogate model is trained on these solutions. Next, search
is performed the same as without a surrogate model, but whenever solutions need
to be evaluated, mainly surrogate (estimated by the surrogate model) evaluations
are performed. Real function evaluations are performed only in case of high
predicted fitness value. The rationale is that solutions with high surrogate
fitness values should also have high real fitness values. However, for the sake
of biasing the search competently toward high-quality solutions, we must be sure
of just exactly how good this solution is.
 
The condition of performing a real function evaluation is defined as follows.
Suppose, $R$ is an array of solutions with known real fitness values and $F$ is
an array of corresponding surrogate fitness values. A real function evaluation
is performed for a solution $s$, if its surrogate fitness $f$ is greater than a
threshold $\mathcal{T}$. This threshold is calculated as a $\lambda$-quantile
value of the sorted values in array $F$. The value of $\lambda$ is dynamically
adjusted. In the beginning, $\lambda=1$, meaning that a real function evaluation
for a solution is performed only if its surrogate fitness exceeds the current
surrogate elitist value. If an algorithm has not found new real elitists for a
while, then we suppose that the current surrogate model is not accurate enough
and it is reasonable to perform real evaluations more elaborately. Therefore,
the value of $\lambda$ is then multiplied by a hyperparameter $\eta$
($0<\eta<1$), effectively relaxing the condition for a real fitness evaluation.
Once an algorithm finds a new elitist, $\lambda$ is reset to 1. After evolving
one solution in SA-P3-GOMEA, the surrogate model is retrained and values of $F$
are re-calculated. The frequency of the model retraining can be potentially
adjusted, but we stick to this natural retraining schedule. Note that the
proposed surrogate-assisted EA has only one numeric hyperparameter $\eta$.

As a surrogate model is unlikely to predict surrogate fitness values with
perfect precision, comparing real fitness values and surrogate fitness values is
undesirable and leads to inferior performance \cite{singh,
dushatskiy2019convolutional}. Therefore, for the comparison of the fitness
values of two solutions the following strategy is used. If both solutions
already have a real fitness value, they are compared via these. Otherwise,
surrogate fitness values are calculated for both solutions and used for the
comparison. The pseudocode of the comparison function is provided in
Algorithm~\ref{alg:P3Func}.

\subsubsection{Surrogate model}
Choosing the best surrogate model is often the key to good performance in
surrogate-assisted optimization. However, the main goal of this paper is not to
propose the best performing surrogate model, but rather to introduce a new
adaptive mechanism for surrogate model integration, use it in P3-GOMEA, and test
its performance on an ensembling problem. To underline the generality of our
proposed mechanism, we experiment with four types of surrogate model.
\begin{enumerate}
  \item Multi-Layer Perceptron (MLP). In contrast to a Convolutional
        Neural Network used in \cite{dushatskiy2019convolutional}, the
        considered Neural Networks are fully connected as we do not want to
        assume any regularity in input data such as subfunctions of fixed size.
        We use a fixed model structure with two hidden layers, each having the
        number of neurons equal to the input size.
  \item Decision Tree-based Gradient Boosting \cite{chen2016xgboost}. This type
        of model was shown to perform well on a variety of tasks
        \cite{zhong2018xgbfemf,chen2018xgboost,dorogush2018catboost} and can
        naturally handle categorical variables.
  \item Support Vector Regression (SVR). This is a common choice for
        surrogate-assisted EAs \cite{surrogate_ea_review}.
  \item Random Forest (RF).
\end{enumerate}
To allow correct processing of categorical variables by MLP, RF and SVR, input
solutions are one-hot encoded. Thus, if an optimization domain has $\ell$
variables with alphabet size $\alpha$, surrogate models get samples of
dimensionality $\ell \alpha$ as input.

Gradient Boosting, RF, and SVR have several hyperparameters such as kernel type
for SVR, and learning rate for Gradient Boosting. Hyperparameters are tuned only
once (before the first model training) using 3-fold cross-validation and simple
grid search. We consider tuning the MLP architecture to be out of the scope of
this paper. The list of tunable hyperparameters and considered values are
provided in the Supplementary, Table 1.

\section{Partition-based ensemble learning}
We consider a well-known supervised classification machine learning problem. The
task is to find a partitioning of the dataset into $K$ disjoint subsets
$C_0,\dots\,C_{K-1}$ ($C_0~\cap~C_1~ \cap~,\dots,~\cap~C_{K-1}=\emptyset$) such
that the aggregated performance of models $M_0,\dots,M_{K-1}$ trained on these
subsets is maximized.

\subsection{Evaluation of fitness} \label{subs:evals}
The aggregated performance is the fitness measure in our EAs. To compute it,
each model predicts class probabilities on the validation dataset (validation
and train datasets do not overlap). Then, for each sample, these probabilities
are averaged and the class with the maximum averaged probability is chosen as
the final prediction. After obtaining predicted classes for all samples, the
final fitness value is calculated as the standard accuracy metric. We use SVM as
learners in constructed ensembles because of its good performance and
computational efficiency (see Section~\ref{sub:ensemble}).

We assume that fitness function evaluations are deterministic, i.e., the
training of SVMs used in an ensemble is deterministic. Moreover, to count only
truly different evaluations (without calculating solutions which define
identical groups of samples twice i.e., 001122 encodes the same partitioning as
112200), all solutions along with their fitness values are stored in a
normalized form. To do so, first, a solution is transformed to an actual
partition: each subset $k$ has $N_k$ samples:
$C_k=\{p_{i_0},\dots,p_{i_{N_{k}-1}}\}$. Then, subsets $C$ are sorted by the
minimal index $p$ of a sample belonging to a subset. A normalized solution form
is then obtained by assigning values to variables according to the sorted order
of subsets. Such solution normalization guarantees that different solutions
defining the same dataset partition are evaluated and counted only once. When a
new solution $s$ is considered to be evaluated, first, it is normalized to $s'$
and looked up in the collection of evaluated solutions. If $s'$ has not been
evaluated, then evaluation is performed, and the obtained fitness value is
stored for $s'$. Note, since we consider the optimization function to be a black
box, the normalized representation is only used to calculate fitness. The
solution itself is left unchanged.

\section{Experimental setup}

\subsection{Datasets}

\begin{table}
\caption{Datasets specification. All datasets are downloaded from the OpenML datasets collection.}
\label{Tab:datasets}
\begin{tabular}{c|c|c|c}
\textbf{Dataset} & \textbf{Samples} & \textbf{Features}& \textbf{Classes}  \\ \hline
segment & 2310 & 16 & 7  \\
spambase & 4601 & 57 & 2  \\
wall-robot-navigation  & 5456 & 24 & 4  \\
kc1  & 2109 & 21 & 2  \\
optdigits & 5500 & 40 & 11  \\
\end{tabular}
\end{table}

We consider five datasets from the OpenML-CC18 benchmark \cite{bischl2017openml}. All datasets are supervised classification tasks and have at least 2000 samples. The selected datasets each have a different number of classes (between 2 and 11), and the number of features ranges from 16 to 57. All features are numeric. Specifications are provided in Table~\ref{Tab:datasets}. From each dataset, 500 randomly selected samples are left out for the validation set, and the same number of randomly selected samples is left out for the test set. The remaining samples are shuffled and $\ell$ random samples are used for a considered optimization problem with $\ell$ variables (i.e., used for training).

\subsection{Ensemble training} \label{sub:ensemble}
We use standard regularized non-linear SVM with Radial Basis Function (RBF) kernel from the Scikit-Learn package. Before training, all features in the datasets are scaled to the unit interval. The regularization parameter of the SVM models is not tuned, but the default value of 1.0 is used. This choice is addressed in the Discussion.

\subsection{Considered optimization algorithms}
As simple baselines, we include in our experiments RS and LS. We also compare
the performance of the proposed SA-P3-GOMEA to P3-GOMEA. Additionally, we
consider three modern BO algorithms. COMBO \cite{combo} has shown
state-of-the-art performance on several combinatorial optimization functions.
Hyperopt \cite{hyperopt} is an implementation of Tree Parzen Estimator (TPE).
SMAC \cite{smac} is a BO algorithm specifically designed to handle categorical
variables by using an RF surrogate model. Finally, we compare the achieved
ensemble performance to training one SVM model on all samples.

\subsection{Problem sizes and runtime budget}
We consider various values of $\ell$ (i.e., the number of samples in the
training dataset), and $\alpha$ (i.e., the number of subsets in a partition),
specifically: $\ell \in \{100,250,500\}$ and $\alpha \in \{2,5,10\}$. On each
dataset in each problem configuration (a combination of $\ell$ and $\alpha$),
all considered search algorithms perform 10 runs. Therefore, there are 50 runs
by each algorithm for each problem instance.

As the global optima of the considered problems are not known, we compare the
algorithms by the convergence trajectory of their averaged performance value.
The averaged performance value after $n$ real function evaluations is calculated
as the elitist fitness function value after that many evaluations, averaged over
all 50 runs.

In our main setup, the fitness function is deterministic and only unique dataset
partitions are evaluated (as described in Section~\ref{subs:evals}). P3-GOMEA,
SA-P3-GOMEA, LS, and RS are run without a time limit until 5000 real
function evaluations are performed. Additionally, we consider the case of noisy
fitness evaluations where ensemble training is stochastic, i.e.,
re-evaluating the same partition results in a different fitness. In that
case, all evaluations are counted.

We found that the considered BO algorithms, especially COMBO and SMAC, are much
slower than P3-GOMEA and SA-P3-GOMEA. We therefore set a time limit of 3 hours
for BO algorithm runs for $\alpha=2,5$ and 5 hours for $\alpha=10$. These values
correspond to the observed upper bounds (rounded to hours) of SA-P3-GOMEA run
time for corresponding problem instances. We found that even for the smallest
considered problem ($\ell=100, \alpha=2$), COMBO could produce $\ell$ randomly
solutions in the intialization phase, but only 3 additional solutions in the
subsequent search phase after 4 hours of running. Hence, we do not include COMBO
in the presented results as within the given time limit it works basically as
RS.

\subsection{Implementation details}
Fitness functions for all conducted experiments are implemented in Python
(version 3.7) with Scikit-Learn, and NumPy packages. All considered search
algorithms use identical fitness functions. Code of
COMBO\footnote{\href{https://github.com/QUVA-Lab/COMBO}{https://github.com/QUVA-Lab/COMBO}},
SMAC\footnote{\href{https://github.com/automl/SMAC3}{https://github.com/automl/SMAC3}},
and
TPE\footnote{\href{https://github.com/hyperopt/hyperopt}{https://github.com/hyperopt/hyperopt}}
algorithms are downloaded from the corresponding repositories. P3-GOMEA,
SA-P3-GOMEA, and LS algorithms are implemented in C++ with calls of Python
fitness functions. All implementations are available at the repository of the
first author\footnote{\href{https://github.com/ArkadiyD/SAGOMEA}{https://github.com/ArkadiyD/SAGOMEA}}. Datasets are downloaded using the OpenML Python API.
Experiments are conducted on a machine with Intel(R) Xeon(R) Silver 4110 CPU.

\section{Results}
\begin{figure*}[h]
    \centering
    \includegraphics[width=0.85\textwidth]{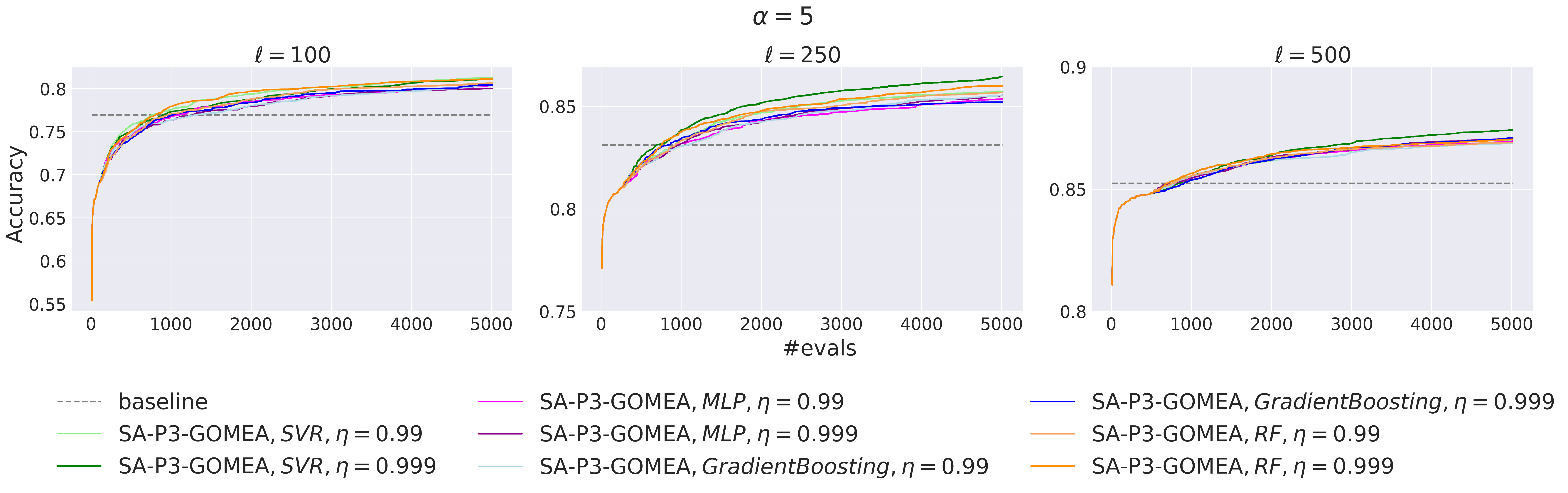}
    \caption{Convergence graphs of average performance for different SA-P3-GOMEA hyperparameters over five datasets and ten optimization runs for each dataset, for $\alpha=5$ (partition into five subsets). Baseline denotes non-ensemble training using all samples in the training dataset, $\ell$ is the number of training samples, i.e., the number of variables in the optimization problem.}
    \label{fig:res1}
\end{figure*}

First, we analyze whether two hyperparameters, the relaxation hyperparameter $\eta$ (described in Section 2.4.1) and the surrogate model type (described in Section 2.4.2), have a significant effect on performance. We test all surrogate models with $\eta=0.99$ and $\eta=0.999$. We do not consider lower $\eta$ values, as they imply a more elaborate use of real fitness evaluations, leading to a performance close to non-surrogate P3-GOMEA. These results are provided in Figure~\ref{fig:res1}. We conclude that 1) while both $\eta$ values provide solid performance, $\eta=0.999$ is slightly preferable and hence, we use it in further experiments; 2) SVR provides the best performance among the considered types of surrogate models. RF is a close runner-up, while MLP and Gradient Boosting models perform worse. We hypothesize that the better performance obtained with SVR is because SVR is less susceptible to overfitting and can generalize well from a limited number of training samples (i.e., extrapolate to unseen solutions). Moreover, it has fewer tunable hyperparameters than, for instance, Gradient Boosting models, and it is feasible to carefully tune them using grid search. Therefore, in further experiments, when we refer to SA-P3-GOMEA, SVR is always used as the surrogate model.

Next, we compare the results of SA-P3-GOMEA to other search algorithms (see Figure~\ref{fig:res2}; Supplementary, Table 2). We observe that P3-GOMEA outperforms LS and RS in most cases. Moreover, the difference becomes larger as the number of variables increases. In general, SA-P3-GOMEA demonstrates solid performance on all problem instances. It always achieves the best accuracy within 500 evaluations, except for the smallest considered problem instance ($\ell=100, \alpha=2$) where after 2000 evaluations SMAC finds better solutions. We observe that SMAC does not scale well to larger problem instances as its computational overhead grows fast when both $\ell$ and $\alpha$ increase. In case of $\alpha=10$ the problem apparently becomes too difficult within the given evaluations limit, as the achieved accuracy is below the baseline for $\ell=100,250$ and the performance of SA-P3-GOMEA is similar to P3-GOMEA. We suppose that in case of $\alpha=10$ the surrogate modelling task is of too high dimensionality ($\ell\alpha$) and therefore, more solutions are needed to get accurate fitness function predictions. However, importantly, due to the adaptive nature of our new surrogate mechanism, even when it is not possible to accurately estimate fitness values, the performance of SA-P3-GOMEA does not become worse than P3-GOMEA.

From the practical perspective of the goal of ensembling, it is further important to note that a better validation accuracy is achieved than the baseline in most cases. Though the accuracy decreases as $\alpha$ increases, we believe it is possible to find better solutions if more function evaluations are allowed.

\subsection{Runtime}
On problem instances with $\alpha=2,5$ the runtime of SA-P3-GOMEA is below 2.5 hours. For $\alpha=10$, runtime does not exceed 4.5 hours.

\subsection{Statistical significance tests}
We employ statistical hypothesis testing to compare the performance of SA-P3-GOMEA with P3-GOMEA and TPE. For each dataset and each value of $\ell$ (5 datasets, 3 values of $\ell$) we do a pairwise statistical test to test a hypothesis that one algorithm performs better than another. Performance is assessed as the best achieved real fitness value during a run. We use the Wilcoxon pairwise test with Holm correction, resulting in a corrected significance level of 0.1. The $p$-values are presented in Table~\ref{tab:tests}. For most problem instances, SA-P3-GOMEA outperforms P3-GOMEA and TPE.

\subsection{Generalization study}
In Machine Learning, a validation dataset is usually used for tuning hyperparameters, model selection, etc. In our setup, we search for the best ensemble configuration based on the accuracy score on the validation dataset. However, to evaluate the generalization ability of the obtained ensembles, we also look at the results on test datasets, which are held out during the search. These results are not related to the evaluation of the optimization potential of the considered algorithms, but are important to assess the practical value of the considered ensembling technique. These results are presented in Supplementary, Figure 1. We observe that convergence curves on test datasets are similar to the ones on validation datasets. Also, SA-P3-GOMEA demonstrates better performance in most cases.

\colorlet{significant}{Blue!30}

\begin{table}
\caption{Statistical significance testing of performance difference in pairs of best-performing algorithms. Reported $p-$values are obtained by pairwise Wilcoxon test and are uncorrected. Statistically significant results at $\alpha=0.1$ with Holm correction ($m=9$) are highlighted.}
\label{tab:tests}
\begin{tabularx}{0.45\textwidth}{c|c|X|X}
     \boldmath${\alpha}$ & \boldmath{$\ell$} & \textbf{SA-P3-GOMEA} is better than \textbf{P3-GOMEA} & \textbf{SA-P3-GOMEA} is better than \textbf{TPE} \\
     \hline 
2 & 100 & 
\cellcolor{significant} 0.025 & \cellcolor{significant} 0.006 \\
2  & 250 & 
\cellcolor{significant} 0.0 &\cellcolor{significant} 0.0 \\
2  & 500 & 
\cellcolor{significant} 0.002 & \cellcolor{significant} 0.006 \\
5  & 100 & 
\cellcolor{significant} 0.002 &\cellcolor{significant} 0.0 \\
5  & 250 & 
\cellcolor{significant} 0.0 &\cellcolor{significant} 0.0 \\
5  & 500 & 
0.069 & \cellcolor{significant} 0.002 \\
10  & 100 & 
\cellcolor{significant} 0.002 &\cellcolor{significant} 0.0 \\
10  & 250 & 
\cellcolor{significant} 0.023 & \cellcolor{significant} 0.0 \\
10  & 500 & 
0.13 & \cellcolor{significant} 0.0 \\

\hline
\end{tabularx}
\end{table}

\section{Discussion and future work}
The main goal of this paper was to introduce a novel efficient
surrogate-assisted EA for solving combinatorial non-binary optimization problems
with expensive function evaluations by integrating a surrogate model into a
P3-like EA through a new adaptive mechanism with only one hyperparameter. Though
we found the value of $\eta=0.999$ for this hyperparameter to work well for the
considered problem, it might need tuning for different expensive optimization
problems.

Below, we identify and discuss interesting research topics for future work.
Through additional experiments, we also considered the case when fitness
evaluations are noisy, i.e., when the same dataset partitioning can result in
different ensemble accuracy scores because learners in ensembles are initialized
with different random seeds. The results for $\alpha=5$ are provided in the
Supplementary, Figure 2. SA-P3-GOMEA performs better than P3-GOMEA for
$\ell=100,250$ and not worse for $\ell=500$. However, we observe that TPE
performs significantly better for $\ell=250,500$. We note that, in contrast to
TPE, neither P3-GOMEA nor SA-P3-GOMEA is specifically designed for solving noisy
optimization problems. Potentially, specialized design choices aimed at solving
noisy problems can improve their performance. However, the achieved accuracy
values are much lower compared to the deterministic fitness function setup. We
therefore conclude, that for practical usage deterministic ensemble training is
preferable. Not only does it make the search problem easier for all considered
algorithms, better solutions with fewer function evaluations can ultimately be
found.

We chose SVM models as learners in the ensemble due to the solid tradeoff of SVM between computational efficiency and good performance. To achieve the best possible predictive accuracy, one may want to use modern Machine Learning models such as Gradient Boosting, or a mixture of models of different types with carefully tuned hyperparameters. Using the proposed approach to achieve state-of-the-art performance on a supervised machine learning problem is an interesting potential future use case.

We did not conduct experiments with more expensive function evaluations, e.g., using deep learning models in an ensemble. As we aimed to study and analyze the performance of the introduced surrogate-assisted EA in multiple setups, including deep learning models was not computationally feasible. However, validating the performance of SA-P3-GOMEA with such learners in an ensemble for specific applications is considered important future research.

Related to this, all considered types of surrogate models are general-purpose regression models and are not designed specifically for the dataset partition problem. We believe that it is possible to further improve the performance of SA-P3-GOMEA if specialized surrogate models, capable of explicitly modelling the task at hand, are designed. Developing such specialized models is however non-trivial and therefore considered a separate research topic.

\section{Conclusion}
We have introduced a novel surrogate-assisted Evolutionary Algorithm for
expensive combinatorial optimization problems based on a state-of-the-art
model-based EA and a novel adaptive surrogate fitness evaluation control
mechanism. The considered model-based EA is a variant of the Gene-pool Optimal
Mixing Evolutionary Algorithm (GOMEA) that uses a Parameterless Population
Pyramid (P3) scheme. To the best of our knowledge, this is the first time a
surrogate model is integrated into a P3 scheme. We have demonstrated that
SA-P3-GOMEA achieves state-of-the-art performance on the problem of partitioning
a dataset for the purpose of ensembling Machine Learning models. We experimented
with different types of surrogate models for fitness values estimation and found
that, among the models considered, SVR provides the best performance. The
proposed SA-P3-GOMEA outperforms non-surrogate-assisted P3-GOMEA, local search,
and various Bayesian Optimization algorithms on various ensemble learning
problems with up to 500 variables of cardinality 10.
 
\begin{acks}
This work is part of the research programme Commit2Data with project number
628.011.012, which is financed by the Dutch Research Council (NWO). We thank the
Maurits en Anna de Kock Foundation for financing a high-performance computing
system.
\end{acks}

\begin{figure*}
    \centering
    \includegraphics[width=0.97\textwidth]{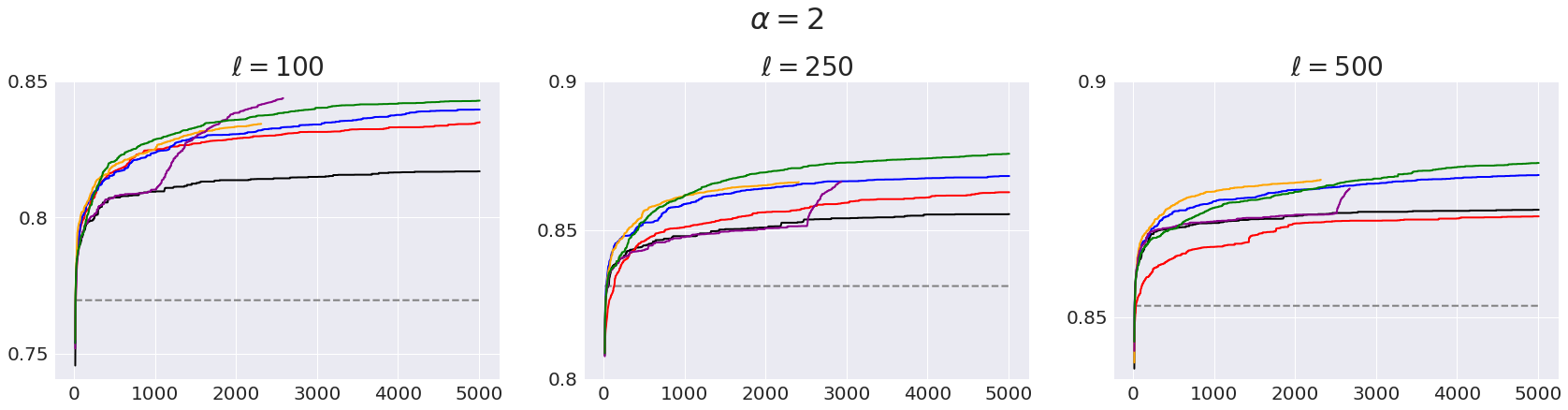}
    \includegraphics[width=0.97\textwidth]{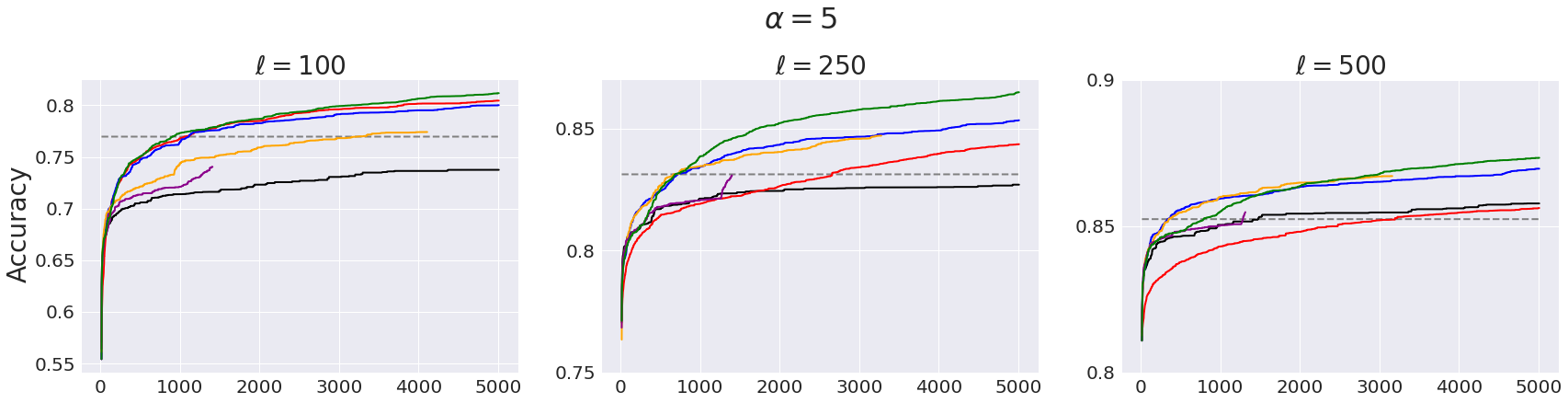}
    \includegraphics[width=0.97\textwidth]{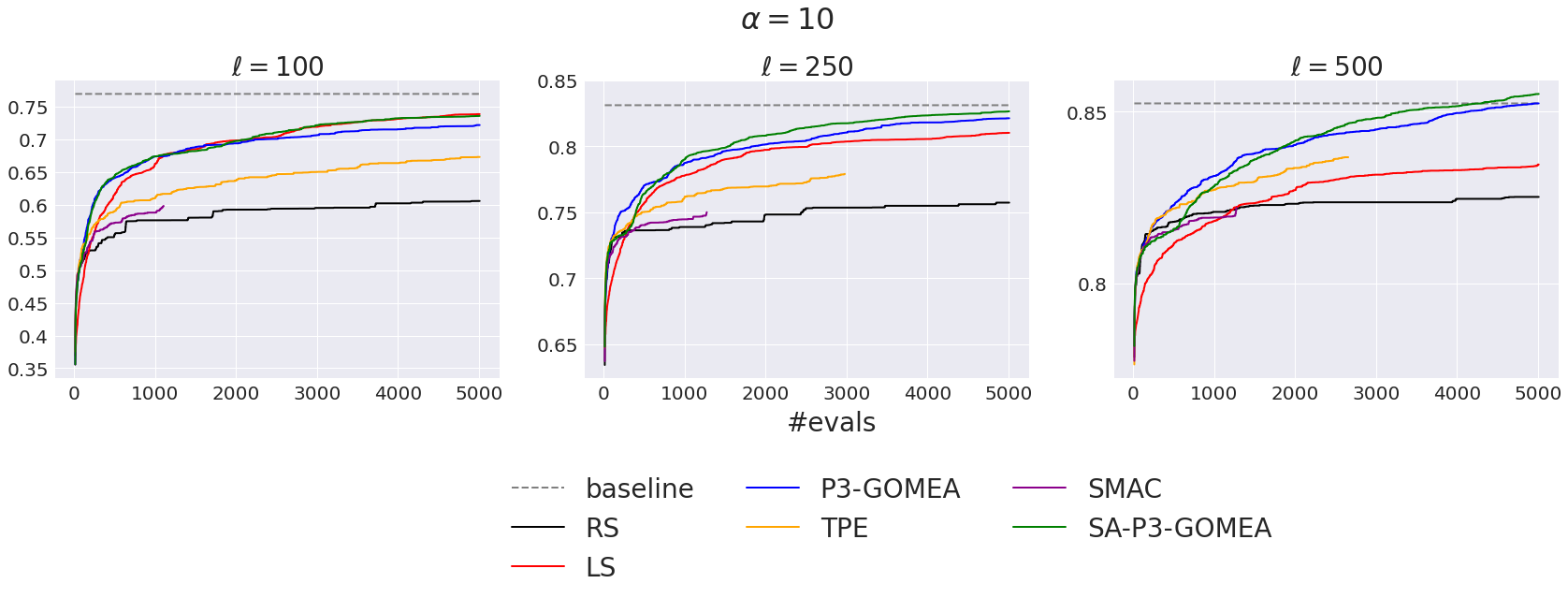}
    \caption{Convergence graphs of average performance for different values of $\alpha$ (the number of partition subsets). $\ell$ denotes the number of training samples, i.e., the number of variables in a corresponding optimization problem. Baseline denotes averaged validation accuracy when SVM is trained on all samples of training datasets of size $\ell$. For TPE and SMAC, the values are presented for fewer than 5000 evaluations as these algorithms could not produce 5000 solutions within the time limit. Note that y-axis scales are different and depend on the accuracy range of the corresponding problem.}
    \label{fig:res2}
\end{figure*}
\clearpage
\onecolumn
\section*{Supplementary}
\setcounter{figure}{0}   
\setcounter{table}{0}   

\begin{table}[h]

\caption{Hyperparameters of considered types of surrogate models. Hyperparameters with multiple listed values are tuned using grid search. SVR model is Support Vector Regression implementation from Scikit-Learn Python library. CatBoostRegressor is Gradient Boosting Regression with categorical variables support from the CatBoost library. MLP is Multilayer Perceptron implementation using PyTorch. RF is Random Forest implementation from Pyrfr (the same library is used in SMAC). Hyperparameters which are not listed in the table are set to default values.}

\begin{tabular}{lll}
    \textbf{model} & \textbf{hyperparameter} & \textbf{values} \\
     \hline 
 \multirow{1}{*}{SVR} & \multicolumn{1}{l}{kernel} & \multicolumn{1}{l}{[rbf, sigmoid]} \\\cline{2-3}
                                 \hline
 
  \multirow{4}{*}{CatBoostRegressor} & \multicolumn{1}{l}{iterations} & \multicolumn{1}{l}{300} \\\cline{2-3}
                                 & \multicolumn{1}{l}{learning\_rate} & \multicolumn{1}{l}{[1.0, 0.1, 0.01]} \\\cline{2-3}
                                 & \multicolumn{1}{l}{depth} & \multicolumn{1}{l}{[3, 6, 9, 12]} \\\cline{2-3}
                                 & \multicolumn{1}{l}{\text{early\_stopping\_rounds}} & \multicolumn{1}{l}{5} \\\cline{2-3}
                                 
                                 \hline
   \multirow{4}{*}{MLP} & \multicolumn{1}{l}{activation\_function} & \multicolumn{1}{l}{ReLU} \\\cline{2-3}
                                 & \multicolumn{1}{l}{Dropout} & \multicolumn{1}{l}{0.2} \\\cline{2-3}
                                 & \multicolumn{1}{l}{n\_hidden\_layers} & \multicolumn{1}{l}{2} \\\cline{2-3} \hline
                                 
    \multirow{4}{*}{RF} & \multicolumn{1}{l}{num\_trees} & \multicolumn{1}{l}{10} \\\cline{2-3}
                                 & \multicolumn{1}{l}{min\_samples\_split} & \multicolumn{1}{l}{[1, 3, 10]} \\\cline{2-3}
                                 & \multicolumn{1}{l}{min\_samples\_leaf} & \multicolumn{1}{l}{[1, 3, 10]} \\\cline{2-3}                               
                                 & \multicolumn{1}{l}{ratio\_features} & \multicolumn{1}{l}{[5/6, 1.0]} \\\cline{2-3} 
                                 
                                 \hline
                                 
\hline
\end{tabular}
\end{table}

\setlength\tabcolsep{1.5pt} 
\begin{table}[h]
    \vspace{3cm}
    \caption{Average accuracy of the elitist solutions (averaged over 10 performed runs) for different algorithms for all datasets and considered problem configurations. The best performing algorithms for each dataset and configuration are highlighted.}
    \label{tab:my_label}
    \centering
    \hspace*{-1cm}
    \begin{tabular}{c|c?c|c|c|c|c|c|c|c|c|c|c|c|c|c|c}
          $\alpha$ & $\ell$ & \multicolumn{3}{c|}{\textit{segment}}& \multicolumn{3}{c|}{\textit  {spambase}}& \multicolumn{3}{c|}{\textit{wall-robot-navigation}}& \multicolumn{3}{c|}{\textit{kc1}}& \multicolumn{3}{c}{\textit{optdigits}} \\ \hline
          & & \footnotesize P3-GOMEA &  \footnotesize TPE &  \footnotesize SA-P3-GOMEA &  \footnotesize P3-GOMEA &  \footnotesize TPE &  \footnotesize SA-P3-GOMEA &  \footnotesize P3-GOMEA &  \footnotesize TPE &  \footnotesize SA-P3-GOMEA &  \footnotesize P3-GOMEA &  \footnotesize TPE &  \footnotesize SA-P3-GOMEA &  \footnotesize P3-GOMEA &  \footnotesize TPE &  \footnotesize SA-P3-GOMEA  \\ \hlinewd{1pt}
\multirow{3}{*}{$2$} & 100 & \textbf{0.843} & 0.842 & 0.83 & 0.904 & 0.9 & \textbf{0.912} & 0.713 & 0.715 & \textbf{0.734} & \textbf{0.865} & \textbf{0.865} & \textbf{0.865} & 0.873 & 0.864 & \textbf{0.874} \\
 & 250 & 0.866 & 0.866 & \textbf{0.868} & 0.918 & 0.916 & \textbf{0.921} & 0.76 & 0.761 & \textbf{0.78} & 0.869 & 0.87 & \textbf{0.872} & 0.929 & 0.928 & \textbf{0.937} \\
 & 500 & 0.857 & \textbf{0.862} & 0.86 & 0.915 & 0.915 & \textbf{0.916} & 0.802 & 0.801 & \textbf{0.805} & 0.869 & 0.871 & \textbf{0.872} & 0.958 & 0.954 & \textbf{0.96} \\
\hline
\multirow{3}{*}{$5$} & 100 & 0.798 & 0.769 & \textbf{0.833} & 0.896 & 0.896 & \textbf{0.897} & 0.697 & 0.691 & \textbf{0.705} & \textbf{0.865} & 0.864 & \textbf{0.865} & 0.745 & 0.653 & \textbf{0.758} \\
 & 250 & 0.861 & 0.852 & \textbf{0.869} & 0.905 & 0.906 & \textbf{0.907} & 0.73 & 0.73 & \textbf{0.76} & 0.867 & 0.868 & \textbf{0.869} & 0.904 & 0.892 & \textbf{0.919} \\
 & 500 & \textbf{0.855} & 0.851 & \textbf{0.855} & \textbf{0.912} & \textbf{0.912} & 0.911 & 0.779 & 0.777 & \textbf{0.795} & 0.869 & \textbf{0.87} & 0.867 & 0.933 & 0.93 & \textbf{0.938} \\
\hline
\multirow{3}{*}{$10$} & 100 & 0.695 & 0.654 & \textbf{0.719} & 0.88 & 0.865 & \textbf{0.884} & 0.659 & 0.629 & \textbf{0.678} & 0.862 & 0.863 & \textbf{0.864} & 0.514 & 0.411 & \textbf{0.535} \\
 & 250 & 0.84 & 0.766 & \textbf{0.85} & \textbf{0.898} & 0.895 & 0.895 & 0.707 & 0.691 & \textbf{0.715} & 0.85 & 0.844 & \textbf{0.86} & 0.811 & 0.714 & \textbf{0.813} \\
 & 500 & \textbf{0.844} & 0.83 & 0.842 & \textbf{0.904} & 0.902 & 0.903 & 0.739 & 0.718 & \textbf{0.748} & \textbf{0.864} & 0.856 & 0.863 & 0.912 & 0.889 & \textbf{0.919} \\
\hline

    \end{tabular}
\end{table}

\begin{figure*}
    \centering
    \includegraphics[width=0.85\textwidth]{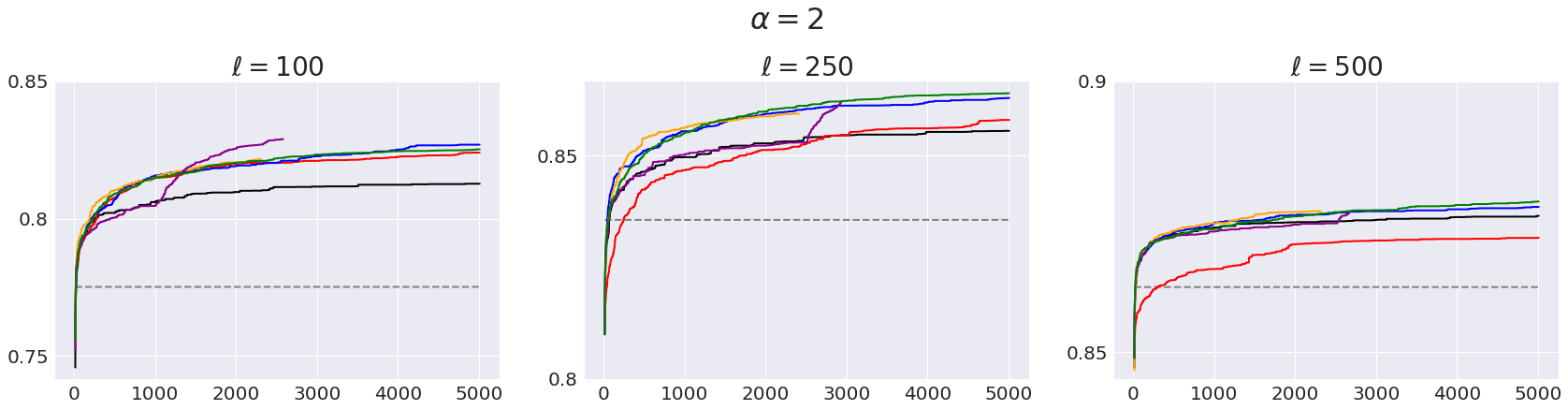}
    \includegraphics[width=0.85\textwidth]{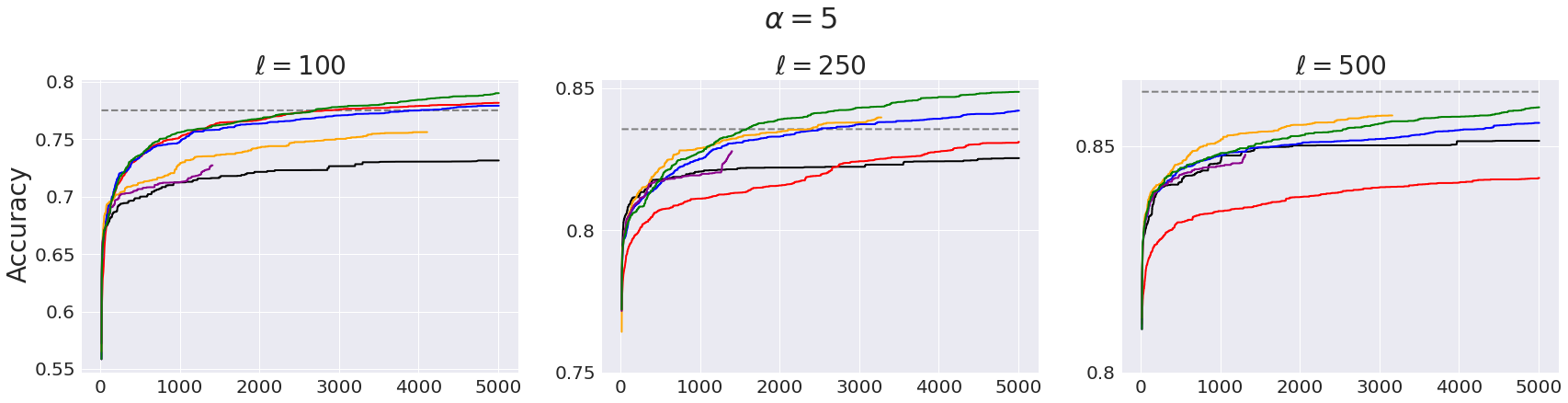}
    \includegraphics[width=0.85\textwidth]{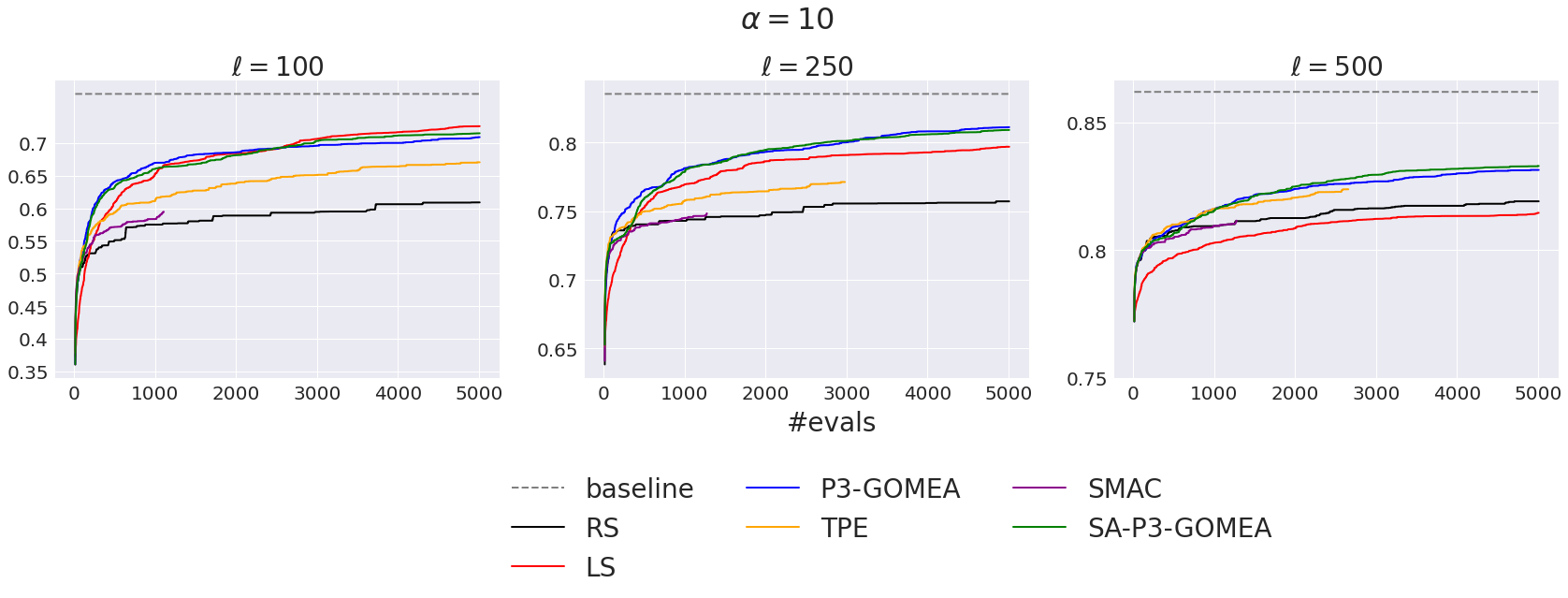}
    
    \caption{Averaged convergence results on \emph{test} dataset for different values of $\alpha$ (the number of partition subsets). $\ell$ denotes the number of training samples, i.e., the number of variables in a corresponding optimization problem. Baseline denotes averaged validation accuracy when SVM is trained on all samples of training datasets of size $\ell$. For TPE and SMAC the values are presented for fewer than 5000 evaluations as these algorithms could not produce 5000 solutions within the given time limit. Note that y-axis scales are different and depend on the accuracy range of the corresponding problem.}
    \label{fig:res2}
\end{figure*}

\begin{figure*}
    \centering
    \includegraphics[width=0.8\textwidth]{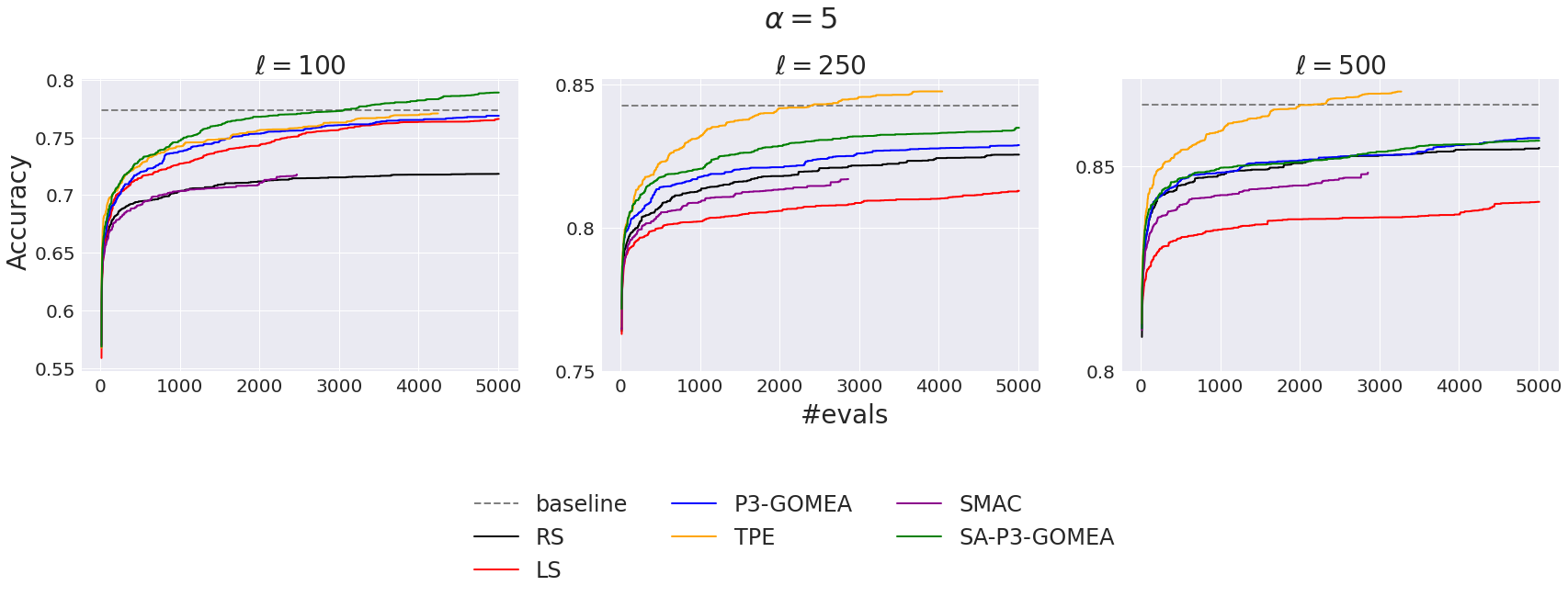}

    \caption{Averaged convergence results for the case of the \emph{noisy} fitness function. $\ell$ denotes the number of training samples, i.e., the number of variables in a corresponding optimization problem. Baseline denotes averaged validation accuracy when SVM is trained on all samples of training datasets of size $\ell$ with 10 different random seeds for each dataset.  SMAC is run with a specific setting for noisy fitness function handling. Note that y-axis scales are different and depend on the accuracy range of the corresponding problem.}
    \label{fig:res3}
\end{figure*}

\clearpage
\bibliographystyle{ACM-Reference-Format}
\bibliography{sample-bibliography} 

\end{document}